\documentclass[11pt,a4paper,onecolumn]{IEEEtran}

\usepackage{algorithmic}
\usepackage{algorithm}
\usepackage[colorlinks=true,citecolor=magenta,urlcolor=blue]{hyperref}
\usepackage{cite}
\usepackage{graphicx,epstopdf}
\usepackage{rotating}
\usepackage{epsfig}
\usepackage[cmex10]{amsmath}
\usepackage{amssymb}
\usepackage{amsthm}
\usepackage{float}
\usepackage[tight,footnotesize]{subfigure}
\usepackage{array}
\usepackage{color}
\usepackage{multirow}
\graphicspath{{images_final/}}
\AppendGraphicsExtensions{.tif}
\usepackage{footnote}
\makesavenoteenv{table}

\ifCLASSINFOpdf
\else
\fi

\begin{document}

\title{Disaggregation of SMAP L3 Brightness Temperatures to 9km using Kernel Machines}
\author{Subit~Chakrabarti,~\IEEEmembership{Student~Member,~IEEE}, ~Jasmeet~Judge,~\IEEEmembership{Senior~Member,~IEEE}, Tara~Bongiovanni, ~Anand~Rangarajan,~\IEEEmembership{Member,~IEEE},\\~Sanjay~Ranka,~\IEEEmembership{Fellow,~IEEE}.%
\thanks{
This work was supported in part by the NASA-Terrestrial Hydrology Program (THP)-NNX13AD04G.

S. Chakrabarti and J. Judge are with the Center for Remote Sensing, 
Agricultural and Biological Engineering Department, 
Institute of Food and Agricultural Sciences, University of Florida, Gainesville, USA; A. Rangarajan and S. Ranka are with the Department of Computer \& Information Science \& Engineering, University of Florida, Gainesville.
E-mail: \href{mailto:subitc@ufl.edu}{subitc@ufl.edu}

A version of this manuscript has been submitted to IEEE Geoscience and Remote Sensing Letters. }}



\maketitle

\begin{abstract}
\boldmath
In this study, a machine learning algorithm is used for disaggregation of SMAP brightness temperatures (T$_{\textrm{B}}$) from 36km to 9km. It uses image segmentation to cluster the study region based on meteorological and land cover similarity, followed by a support vector machine based regression that computes the value of the disaggregated T$_{\textrm{B}}$ at all pixels. High resolution remote sensing products such as land surface temperature, normalized difference vegetation index, enhanced vegetation index, precipitation, soil texture, and land-cover were used for disaggregation. The algorithm was implemented in Iowa, United States, from April to July 2015, and compared with the SMAP L3\_SM\_AP T$_{\textrm{B}}$ product at 9km. It was found that the disaggregated T$_{\textrm{B}}$ were very similar to the SMAP-T$_{\textrm{B}}$ product, even for vegetated areas with a mean difference $\leq$ 5K. However, the standard deviation of the disaggregation was lower by 7K than that of the AP product. The probability density functions of the disaggregated T$_{\textrm{B}}$ were similar to the SMAP-T$_{\textrm{B}}$. The results indicate that this algorithm may be used for disaggregating T$_{\textrm{B}}$ using complex non-linear correlations on a grid.
\end{abstract}



\newpage
\section{Introduction}
Microwave observations are highly sensitive to near-surface soil moisture (SM)~\cite{Jackson1991,Schmugge1995,Ahmad2010,Gruhier2010,Barrett2009,Qin2009,Lakhankar2009,Mao2008,Wang2007,Dongsheng2007}, and with the recent launch of the NASA Soil Moisture Active and Passive (SMAP) mission along with the ESA SMOS mission, we now have unprecedented global observations of SM every 2-3 days. The NASA-SMAP was expected to provide SM at a spatial resolution of 9~km retrieved from combined active and passive (AP) observations at 1.26 and 1.41 GHz, respectively, every 2-3 days~\cite{entekhabi2008}. This spatial resolution is optimal for many hydrometeorological applications~\cite{Fennessy2009,Douville2000,Koster2004,Tubiello2002,Yuste2005,Friend2007,Fecan1999}. However, on July 7\textsuperscript{th} 2015, the radar on board SMAP halted its transmissions due to an anomaly~\cite{SMAPWeb}, creating a gap in the disaggregation of coarse scale radiometer observations (T$_{\textrm{B}}$) available at 36km to 9km using active observations to meet the mission requirements for the L3\_SM\_AP product. A few studies have attempted to disaggregate T$_{\textrm{B}}$ directly without the complementary information provided by active observations, such as statistical inversion techniques~\cite{YongQian2011,Gambardella2008,Lenti2014}. These techniques allow discovery of non-linear correlations between T$_{\textrm{B}}$ across scales. However, many of these methods require high-resolution T$_{\textrm{B}}$ for training, which is typically not available. Piles et. al.~\cite{Piles2012} disaggregated T$_{\textrm{B}}$ directly into SM by applying the Universal Triangle (UT) method and used a 2$^\mathrm{nd}$ order regression-based linking model to relate coarse resolution SM to T$_{\textrm{B}}$ from the SMOS mission, and other high resolution products, aggregated to the resolution of SMOS observations.  The fine scale SM was then estimated using the assumption that the linking model at the coarse resolution also holds at finer resolutions. The robustness of this method over heterogeneous vegetation and weather conditions remain mostly untested. Treating each pixel as a sample instead of using spatial information to regularize the disaggregation results in salt and pepper noise due to spatial auto-correlation\cite{Jiang2013}. Moreover, these approaches use second order metrics, which do not leverage all the information in the data that is necessary in a highly non-linear regression problem such as disaggregation\cite{Principe2010}.

The goal of this study is to disaggregate SMAP T$_{\textrm{B}}$ observations from 36~km to 9~km without requiring \emph{in-situ} data for training. This is done using a version of the self regularized regressive model~(SRRM) algorithm~\cite{Chakrabarti2015} that includes kernel based support vector regression models in conjunction with coarse-scale spatial segmentation. The segmentation algorithm separates the study region into discrete sets of pixels which have similar terrain conditions. A support vector regression model is estimated for each set using coarse scale T$_{\textrm{B}}$ and auxiliary correlated data which is further applied to the auxiliary data at fine scale to obtain disaggregated T$_{\textrm{B}}$. The primary objectives are to, 1) enhance the SRRM algorithm to disaggregate coarse SMAP $\mathrm{T_B}$ to 9~km without fine-scale T$_{\textrm{B}}$ as training, and using other spatially correlated variables such as land surface temperature(LST), normalized difference vegetation index (NDVI), enhanced vegetation index (EVI), precipitation (PPT), soil texture and land-cover (LC), 2) implement the downscaling algorithm in Iowa in mid-western United States, and 3) compare the disaggregated T$_{\textrm{B}}$ to the SMAP L3\_SM\_AP T$_{\textrm{B}}$ at 9km product (hereafter referred to as SMAP-$\mathrm{T_B}$) during April 27 to July 5, 2015, when the product is available.

\section{Theory}
\begin{figure}[t]
\centering\noindent
\centering\noindent\includegraphics[width=7 in, trim = 15 150 15 40,clip]{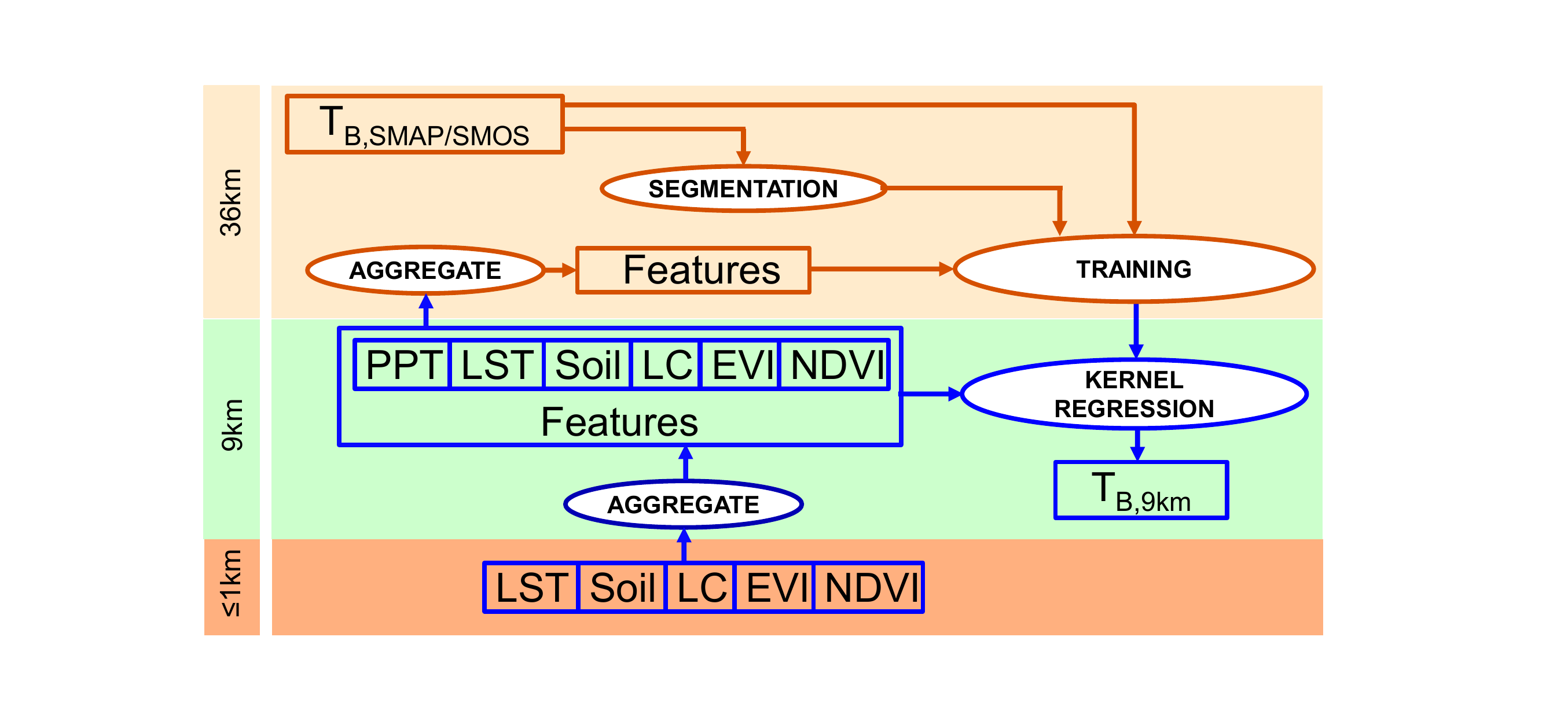}
\caption{Flowchart for the multi-scale SRRM disaggregation method.}
\label{fig:flow}
\end{figure}

Disaggregation is an ill-posed problem constrained by the smoothness of the coarse-scale data which constrains the generation of data at fine scale because any added sharpness can be misconstrued as noise. Additional geospatial and meteorological data that are correlated to T$_{\textrm{B}}$ are needed to ensure that the added sharpness has physical basis. In this study, these auxiliary data-sets are utilized to create localized regression models that provide a mapping from the T$_{\textrm{B}}$ at coarse scale to T$_{\textrm{B}}$ at fine scale. To ensure that the localization is realistic and not arbitrary, the  coarse T$_{\textrm{B}}$ image is first segmented into multiple regions of radiometric similarity. The overall organization and the datasets involved is shown in Figure~\ref{fig:flow}.

\subsection{Self-Regularized Regressive Models (SRRM)}
\label{Sec:OurModel}

The \emph{first step} of the algorithm divides the study area into segments using the coarse scale T$_{\textrm{B}}$. In this study, the segmentation algorithm uses information theoretic measures of inter and intra segment similarity~\cite{Jenssen2005}. If $\mathbf{X}=\{\mathbf{x}_1,\mathbf{x}_2,\mathbf{x}_3\ldots\mathbf{x}_N\}$ is a matrix containing T$_{\textrm{B}}$ for $N$ pixels, the Cauchy-Schwarz cost-function, $\hat{J}_{CS}$, estimates optimal memberships of the pixels to segments, $\mathbf{m}$, in an un-supervised manner. 
\begin{align}
\label{eq:JCS_estimator_reg}
\hat{J}_{CS}^{REG} &= \frac{\frac{1}{2} \sum_{i=1}^{N} \sum_{j=1}^{N} \left(1 - \textbf{m}_i^\mathrm{T} \textbf{m}_j\right) G_{\sigma \sqrt{2}} \left(\mathbf{x}_i,\mathbf{x}_j\right)}{\sqrt{\prod_{k=1}^{K} \sum_{i=1}^{N} \sum_{j=1}^{N} m_{ik}m_{jk}G_{\sigma \sqrt{2}}(\mathbf{x}_i,\mathbf{x}_j)}}
\end{align}
where, $K$ is the number of segments, $G_{\sigma\sqrt{2}}$ is the Gaussian kernel with standard-deviation $\sigma$ and $\nu$ is the regularization weight.
The optimal value of the membership vector can be obtained from the following constrained optimization problem,
\begin{align}
\label{eq:optimization}
&\text{min}_{\substack{\mathbf{m}_1,\dots,\mathbf{m}_N}} \hat{J}_{CS}^{REG}(\mathbf{m}_1,\dots,\mathbf{m}_N) \quad \nonumber \\ &\text{subject to } \mathbf{m}_j^\mathrm{T}\mathbf{1} - \mathbf{1} = 0, \quad j = 1,\dots ,N 
\end{align}

To compute optimum values of $\mathbf{m}$, and thus the membership of each pixel to the $K$ segments, a Lagrange multiplier formulation can be used along with a stochastic gradient descent scheme, the details of which are shown in \cite{Chakrabarti2015}. 

In the \emph{second step}, support vector regression (SVR)~{\cite{vapnik1996}}, is used to generate the downscaled estimates. A training set of pixels, for example $y_{train}$ is used in the regression to fit a non-linear function, for example $f$, from the set of the auxiliary data and coarse scale T$_\mathrm{B}$, for example $\mathbf{z}$, to fine scale T$_\mathrm{B}$. This function takes the form,

\begin{align}
f(z) = <w,z>_{\mathcal{H}} + b
\end{align}

where $w$ are the weights and $<.,.>$ is the inner-product operation in some Hilbert space, $\mathcal{H}$. The cost function of support vector regression, which minimizes the errors between $y_{train}$ and $f(z)$ to at most $\epsilon$, 

\begin{align}
\label{eq:costsvm}
&\frac{1}{2}\|w^2\| + C\sum_{i=1}^l(\xi_i+\xi^*_i) \\
\text{subject to }& 
\begin{cases}
y_{train}-<w,z>_{\mathcal{H}} - b \leq \epsilon+\xi_i \nonumber \\
<w,z>_{\mathcal{H}} + b -  y_{train} \leq \epsilon+\xi^*_i \nonumber \\
\xi^*_i,\xi_i \geq 0
\end{cases}
\end{align}

where $\xi^*_i$ and $\xi_i$ are called \emph{slackness constants} such that Equation~\ref{eq:costsvm} can be solved using convex optimization and $C$ determines the trade-off between the flatness of $f$ and the amount up to which deviations greater that $\epsilon$ are tolerated. The function $f$ is allowed to be non-linear by selecting a suitable $\mathcal{H}$ such that the inner product becomes a kernel evaluation as $<w,z>_{\mathcal{H}} = \kappa(w,z)$. In this study the radial basis function kernel was chosen, $\kappa(w,z) = \mathrm{exp}\left(\frac{\|w-z\|^2}{2\sigma^2}\right)$ where $\sigma$ is the kernel parameters. More details about the statistics and convex optimization theory that is used to solve SVR based problems are available in \cite{smola2004,clarke2004} and are not repeated here.

\section{Experimental Description and Results}
\subsection{Study Area}

\begin{figure}[t]
\centering\noindent
\includegraphics[width=5 in ,keepaspectratio=true,trim = 0 20 0 10,clip]{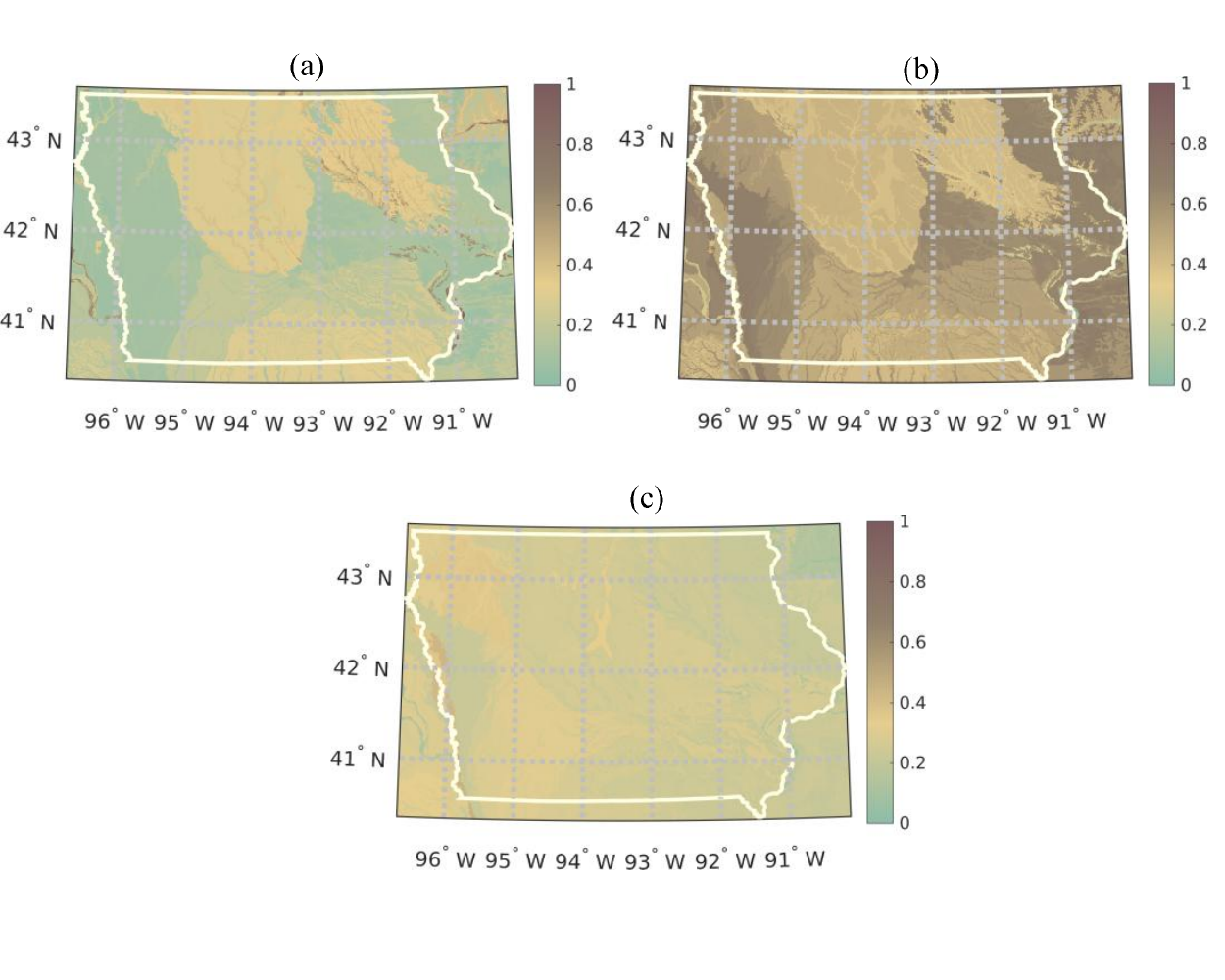}
\caption{Soil texture in the study region in Iowa, United States - volumetric ratio of (a)sand, (b)clay and (c)silt in the soil.}
\label{fig:soil}
\end{figure}

The study was conducted in a $320 \times 560\text{ }\mathrm{km^2}$ region in the state of Iowa in the United States (US), stretching from $40.36^\circ$ to $43.57^\circ \mathrm{ N}$ and $90.14^\circ$ to $96.68^\circ \mathrm{ W}$, equivalent to 162 SMAP pixels.  Iowa, with an area of about $1.7\times10^5$ $\mathrm{km^2}$, out of which $1.2\times10^5$ $\mathrm{km^2}$ is cropland, is one of the most important agricultural areas in the US responsible for $> 70$\% of the country's agricultural gross domestic product~\cite{usda2012}, and includes a SMAP core validation site in the South Fork watershed. The percentage of silt, clay and soil is also shown in the study region in Figure~\ref{fig:soil}.

\begin{figure}[t]
\centering\noindent
\includegraphics[width=5 in ,keepaspectratio=true, trim = 0 0 0 0,clip]{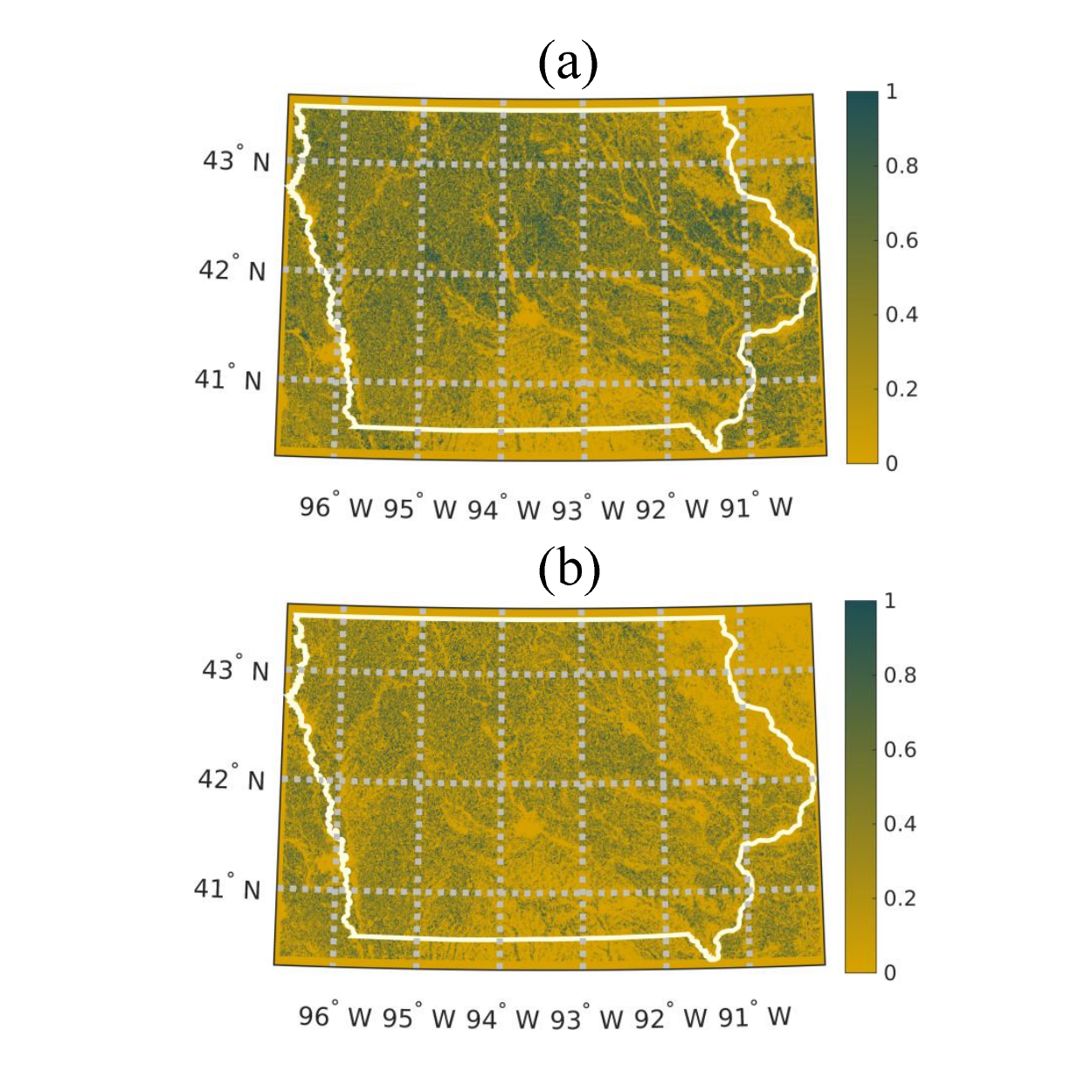}
\caption{Land Covers in the study region in Iowa, United States - (a) Corn, (b) Soybean.}
\label{fig:studysite}
\end{figure}

To downscale T$_{\textrm{B}}$ at 36 km to 9 km, satellite based observations of EVI, NDVI, LST, PPT were used from April 20 to June 30 2015. LC and soil texture (STEX) were also used and considered to be constant for the duration of the study. The products used in this study along with their native and interpolated spatio-temporal resolutions are listed in Table \ref{table:sat}. LST, EVI and NDVI, available at $\sim 1$~km, were interpolated to 9km through spatial averaging. For this study, LC is available as a categorical data product at a spatial resolution of 30~m. LC was recategorized into seven groups and the ratio of each group within a 1~km pixel was used. The two groups occupying the largest area had corn and soybean land covers. Agricultural areas with other crops were conglomerated and referred to as 'miscellaneous' and used as a single group. Some forests, wetlands, and developed regions were also present in the study area and used as three groups. Anything that could not be categorized into the 6 groups, mentioned above, were assigned to a group referred to as 'others'. The ratio of each group within a single 1-km pixel is shown in Figures~\ref{fig:studysite}(a) through (f). The 9~km LC was obtained by linearly averaging the ratios obtained at 1~km. 

\subsection{Implementation in Iowa}

In the modified SRRM algorithm, the $D_{CS}$ based clustering algorithm to discover regions of similarity in the study area. In Equation~\ref{eq:JCS_estimator_reg} from Section II.A., $\mathbf{X}=\{\left[\mathrm{T_{B,1}},\mathrm{lat}_1,\mathrm{lon}_1\right],\left[\mathrm{T_{B,1}},\mathrm{lat}_2,\mathrm{lon}_2\right],\dots,\left[\mathrm{T_{B,M}},\mathrm{lat}_M,\mathrm{lon}_M\right]\}$ where $M$ is the total number of coarse pixels in the region and $\mathrm{lat_i}$ and $\mathrm{lon_i}$ are the latitude and longitude of the $i^{th}$ pixel. The number of clusters, $N$, is used as a parameter. Since no ground truth is available in the region, $N$ is determined using the principle of minimum description length as described in ~\cite{hanse2001} is used. After clustering, $\mathrm{M}$ models, $\hat{f}_1, \hat{f}_2,\dots,\hat{f}_M$ are developed using NDVI, EVI, PPT, LST, LC and STEX aggregated to 36km along with the coarse scale $\mathrm{T_B}$ using Equation~\ref{eq:costsvm}. The disaggregated value of $\mathrm{T_B}$ at 9km is then computed by applying the learnt functions $\hat{f}_1, \hat{f}_2,\dots,\hat{f}_M$ to NDVI, EVI, PPT, LST and LC values aggregated to 9km.

\begin{figure}[t]
\centering\noindent
\includegraphics[width=6.5 in ,keepaspectratio=true]{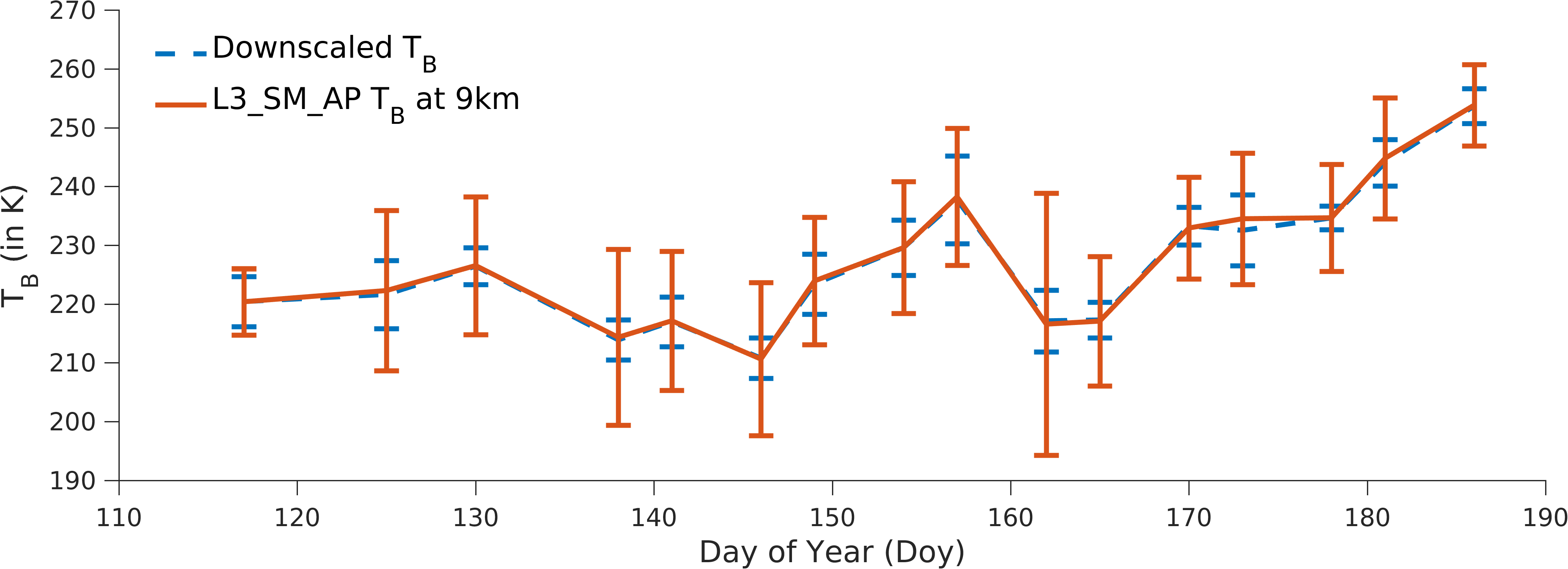}
\caption{Spatial average and variance of disaggregated brightness temperature and SMAP L3\_SM\_AP product at 9km for the whole season.}
\label{fig:RMSE}
\end{figure}

The means and standard deviations of the disaggregated $\mathrm{T_B}$ and SMAP-$\mathrm{T_B}$ are compared to provide an index of the intra-season variability captured in the $\mathrm{T_B}$, as shown in Figure~\ref{fig:RMSE}. The means are preserved by the multiscale SRRM algorithm with differences of $\leq$ 0.5K, while the average standard-deviation of the disaggregated $\mathrm{T_B}$ is 7K lower than that for the SMAP-$\mathrm{T_B}$. The maximum difference between the standard deviations of the disaggregated $\mathrm{T_B}$ and SMAP-$\mathrm{T_B}$ is observed when $\mathrm{T_B}$ shows a decreases sharply on DoY 139 and DoY 161 due to high spatial variability in PPT, and the $\mathrm{T_B}$ is underestimated for areas with high PPT which decreases the standard deviation in disaggregated $\mathrm{T_B}$ because of a 4-hour time lag between the SMAP and Global Precipitation Measurement (GPM) observations. Any rainfall that occurs in this time is not indicated by the GPM PPT product but affects the $\mathrm{T_B}$.

\begin{figure}[t]
\centering\noindent
\includegraphics[width=5 in ,keepaspectratio=true, trim = 15 70 0 20,clip]{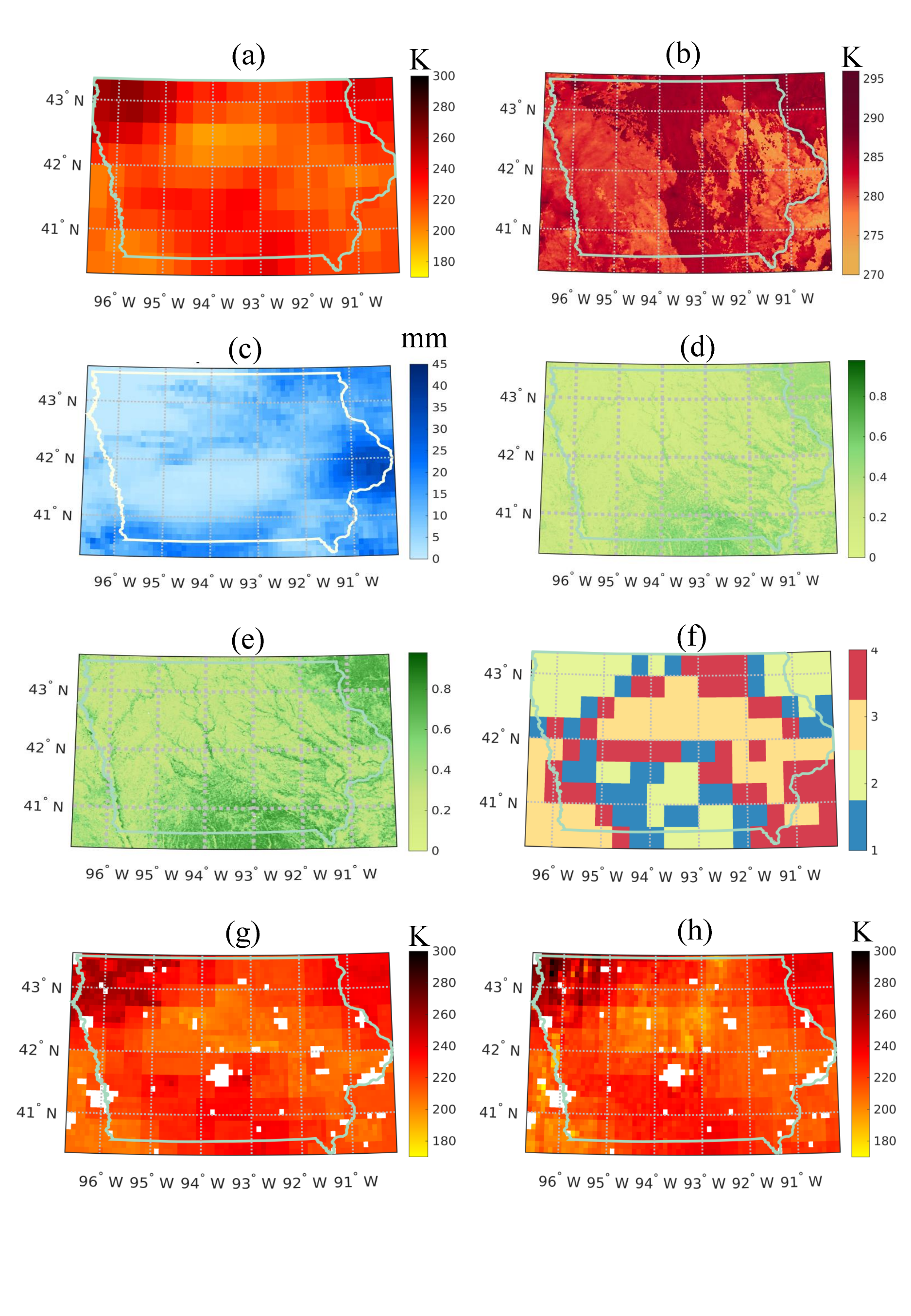}
\caption{(a) SMAP brightness temperature product at 36km, (b) LST at 1km, (c) Precipitation at 9km, (d) Normalized Difference Vegetation Index at 1km, (e) Enhanced vegetation index at 1km, (f) Segmentation at 36km, (g) Disaggregated brightness temperature at 9km, (h) SMAP L3\_SM\_AP product at 9km on May 25, 2015 (DOY 125)}
\label{fig:125}
\end{figure}

\begin{figure}[t]
\centering\noindent
\includegraphics[width=5 in ,keepaspectratio=true, trim = 15 70 10 20,clip]{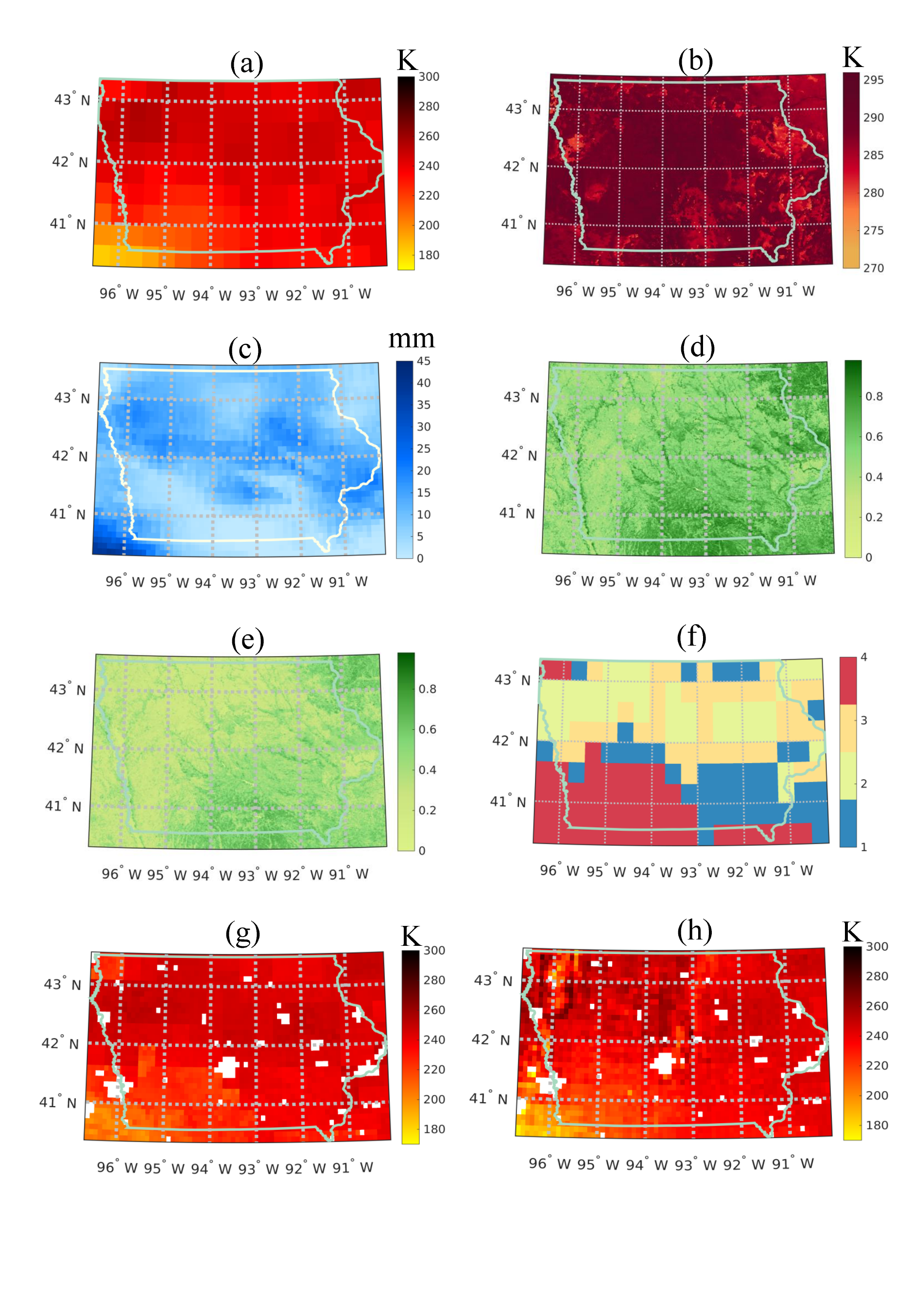}
\caption{(a) SMAP brightness temperature product at 36km, (b) LST at 1km, (c) Precipitation at 9km, (d) Normalized Difference Vegetation Index at 1km, (e) Enhanced vegetation index at 1km, (f) Segmentation at 36km, (g) Disaggregated brightness temperature at 9km, (h) SMAP L3\_SM\_AP product at 9km on June 6, 2015 (DOY 157)}
\label{fig:157}
\end{figure}

\begin{figure}[t]
\centering\noindent
\includegraphics[width=5 in ,keepaspectratio=true, trim = 0 70 15 20,clip]{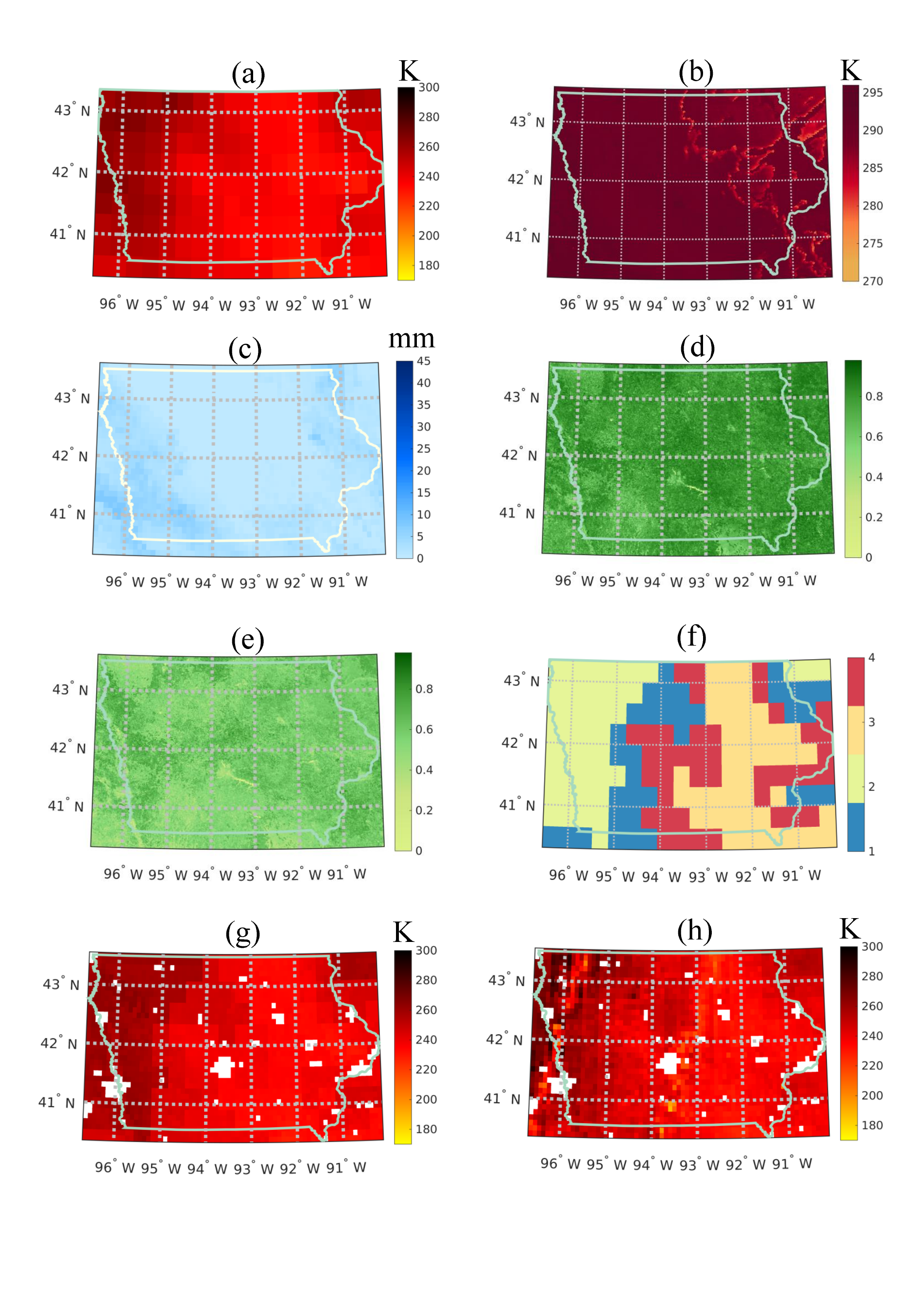}
\caption{(a) SMAP brightness temperature product at 36km, (b) LST at 1km, (c) Precipitation at 9km, (d) Normalized Difference Vegetation Index at 1km, (e) Enhanced vegetation index at 1km, (f) Segmentation at 36km, (g) Disaggregated brightness temperature at 9km, (h) SMAP L3\_SM\_AP product at 9km on June 30, 2015 (DOY 185)}
\label{fig:181}
\end{figure}

In addition, the disaggregated $\mathrm{T_B}$ images on three days during May and June are compared with SMAP-$\mathrm{T_B}$, to determine spatial diversity under increasing amounts of vegetation and different precipitation conditions. On DoY 125 (May 5), the precipitation is heterogeneous across the region, as shown in Figure~\ref{fig:125}(b). The disaggregated $\mathrm{T_B}$ captures the decrease in $\mathrm{T_B}$ in the east-central region of Iowa due to high PPT adequately, similar to the SMAP-$\mathrm{T_B}$ at 9km. The major variability in the inputs and SMAP-$\mathrm{T_B}$ are present in the disaggregated $\mathrm{T_B}$. It is also smoother compared to SMAP-$\mathrm{T_B}$ which suggests low noise levels because any two neighbouring pixels differ by $\leq$ 10K in the disaggregated $\mathrm{T_B}$.  Similar performance is observed for DoYs 157 (June 6) and 181 (June 30), shown in Figure~\ref{fig:157} and Figure~\ref{fig:181} respectively. The locations of the clusters also change on both the days according to the spatial patterns of $\mathrm{T_B}$. Furthermore, even when the NDVI and EVI are high, as observed for DoY 181, the disaggregated $\mathrm{T_B}$ is similar as compared to the SMAP-$\mathrm{T_B}$. On DoY 181, the LST is unavailable for a lot of the pixels in the study region, which reduces the heterogeneity in disaggregated $\mathrm{T_B}$ because the spatial patterns in LST cannot be utilized for disaggregation. However, as shown in Figure~\ref{fig:181}(b), the spatial distribution of the disaggregated $\mathrm{T_B}$ is comparable to the SMAP-$\mathrm{T_B}$. Thus,the multiscale SRRM algorithm is sufficiently robust to vegetation levels and LCs.

\begin{figure}[t]
\centering\noindent
\includegraphics[width=7 in ,keepaspectratio=true]{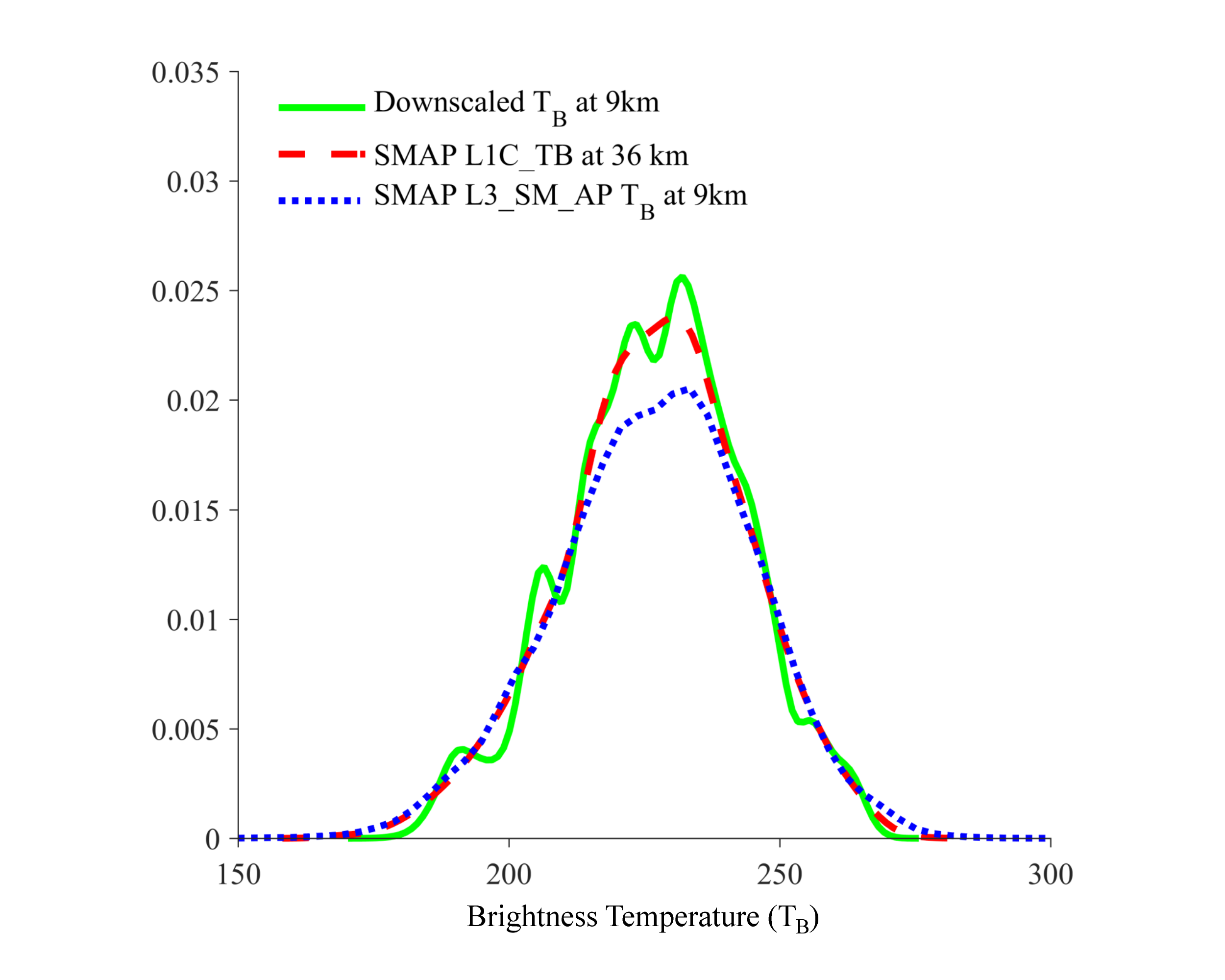}
\caption{Kernel density estimate of the probability density function (PDF) of SMAP L1C\_TB product at 36km, SMAP L3\_SM\_AP product at 9km and Disaggregated brightness temperature at 9km.}
\label{fig:pdf}
\end{figure}

The probability density functions (PDF) of the disaggregated $\mathrm{T_B}$, coarse scale SMAP $\mathrm{T_B}$ at 36km, and SMAP-$\mathrm{T_B}$ at 9km are estimated to facilitate inter-comparisons and further elucidate the relationship among the three products.  The statistical similarity of disaggregated $\mathrm{T_B}$ to the coarse scale SMAP $\mathrm{T_B}$ is shown in Figure~\ref{fig:pdf}. The PDF of disaggregated $\mathrm{T_B}$ is closer to the PDF of coarse scale SMAP $\mathrm{T_B}$ than the SMAP-$\mathrm{T_B}$ which shows that the disaggregated $\mathrm{T_B}$ is more closely coupled to the coarse scale SMAP $\mathrm{T_B}$. This demonstrates that the multi-scale SRRM algorithm can be operationally used to disaggregate the coarse scale SMAP $\mathrm{T_B}$ to 9~km.

\section{Conclusion}

In this study, a disaggregation methodology from 36km to 9km was developed and implemented that preserves the high variability in $\mathrm{T_B}$ due to heterogeneous meteorological and vegetation conditions. The multiscale SRRM preserves heterogeneity by utilizing a segmentation algorithm to create a number of regions of similarity which subsequently, are used in a support vector machine regression framework. The clusters were computed using RS products, \textit{viz.} PPT,  EVI, NDVI, LC and Soil Texture. It was found that the difference between the means of the disaggregated $\mathrm{T_B}$ and SMAP-$\mathrm{T_B}$ is $\leq$ 5K for all days while the average variance is $\leq$ 7K. The disaggregated $\mathrm{T_B}$ and SMAP-$\mathrm{T_B}$ were alike even under highly vegetated conditions. The PDFs of the disaggregated $\mathrm{T_B}$ were found to be closer to the PDF of the coarse scale SMAP $\mathrm{T_B}$ than the PDF of the SMAP-$\mathrm{T_B}$ product. The results indicate that this algorithm can be used for disaggregating T$_{\textrm{B}}$ with complex non-linear correlations on a grid with high accuracy.

\renewcommand{\baselinestretch}{1}
\footnotesize
\bibliographystyle{IEEEtran}
\bibliography{combined}

\clearpage
\begin{table}
\renewcommand{\arraystretch}{1.5} 
\centering
\caption{Satellite observations and other data products used.}
\label{table:sat}
\begin{tabular}{c|m{5.5cm}|c|m{3cm}|c}
\hline
\multicolumn{1}{c|}{Sl. No.}&\multicolumn{1}{c|}{Physical Quantity}&\multicolumn{1}{p{1.5cm}|}{Spatial Resolution}&\multicolumn{1}{c|}{Source}&\multicolumn{1}{p{1.5cm}}{Temporal Resolution}\\
\hline
1&Brightness Temperature&36 km&NASA-SMAP&2-3 days\\
2&Precipitation&0.1\textsuperscript{$\circ$}&NASA-GPM&30 min\\
3&Enhanced Vegetation Index &1 km&NASA Aqua/Terra&8 days\\
4&Land Surface Temperature&1 km&NASA Aqua/Terra&1 day\\
5&Land Cover&30 m&USDA NASS-CDL&1 year\\
6&Normalized Difference Vegetation Index&0.125\textsuperscript{$\circ$}&NASA Aqua/Terra&8 days\\
\hline        
\end{tabular}
\end{table}

\clearpage

\clearpage

\clearpage

\clearpage

\clearpage

\clearpage

\clearpage

\clearpage

\end{document}